\documentclass[11pt]{article}

\usepackage[T1]{fontenc}
\usepackage[margin=1in]{geometry}
\usepackage{amsmath,amssymb}
\usepackage{booktabs}
\usepackage{hyperref}
\usepackage{xcolor}
\usepackage{microtype}
\usepackage{graphicx}
\usepackage{cite}

\title{Routing Absorption in Sparse Attention:\\
Why Random Gates Are Hard to Beat}
\author{Keston Aquino-Michaels\\No Way Labs}
\date{February 2026}

\begin{document}
\maketitle

\begin{abstract}
Learned gates can approximate sparse attention patterns on frozen transformers,
yet provide limited benefit over random gates when trained jointly with the
model. We investigate this difference in a controlled 31M-parameter transformer
and attribute it to \emph{routing absorption}: model representations co-adapt to
the imposed mask, reducing the incremental benefit of learned routing. Four
experiments characterize the phenomenon. Differentiable soft gating yields
perplexities of $48.73 \pm 0.60$ with learned gates and $49.83 \pm 0.04$ with
frozen random gates over three seeds. Hard top-$k$ masking provides no gradient
path to the gate scores in the tested implementation. Gates distilled onto
co-adapted and dense-trained Q/K/V both achieve high F1 against oracle masks,
but hard-mask deployment yields perplexities of 601.6 and 48.6, respectively.
Stochastic mask training also leaves a substantial deployment penalty: dense
evaluation yields 78.2 perplexity, compared with 37.3 for the dense baseline.
Experiments on Qwen3-1.7B show that increasing the number of trainable attention
layers reduces the gap between learned and random gates. We relate these results
to co-adaptation in Mixture-of-Experts and propose parameter asymmetry between
the gate and the model as a contributing mechanism. For the tested per-query,
token-level gates, freezing the model provides stable routing targets and enables
effective post-hoc sparsification. The results motivate random-routing controls
and separate evaluation of routing quality and model adaptation in sparse
attention methods.
\end{abstract}

\section{Introduction}

Dense attention distributions in the models studied here are highly concentrated.
In a 31M-parameter transformer trained on WikiText-103, the top 64
out of 512 key positions per query capture 90.6\% of the total
attention mass~(Section~\ref{sec:structure}).
In Qwen3-1.7B, the concentration is even sharper: oracle top-$k$
masking at $k$=64 (87.5\% sparsity) raises perplexity from 11.52
to 11.57. These measurements motivate learning a compact gate to select
attention entries while preserving model quality.

Post-hoc distillation achieves this in the tested models.
A lightweight bilinear gate ($d_\text{gate}$=32, 1.3\% of model
parameters) trained post-hoc on a frozen dense checkpoint converges
to near-oracle routing in 1{,}000 steps, closing $>$94\% of the gap
between random and oracle masks at all tested sparsity
levels~(Table~\ref{tab:posthoc}).
The same gate architecture, trained end-to-end alongside the model's
Q/K/V projections for 50{,}000 steps, provides a smaller benefit:
its perplexity is 2.2\% lower than that of a frozen random gate.

We investigate how joint training changes the contribution of the gate.
Our results support \emph{routing absorption}, in which model adaptation to the
mask reduces the benefit of learning the routing function. The model's
parameters ($\sim$31M) outnumber the gate's by approximately 80:1 and can adjust
to the imposed masking function. After 50{,}000 steps of co-training, removing
the soft gate or replacing its scores with random noise has little effect, and
the gate's predictions contain little information about attention structure
beyond chance.
Related findings in Mixture-of-Experts (MoE) show that expert co-adaptation
can reduce the difference between random and learned
routing~\cite{roller2021hash,chen2023smoe,li2024empiricalmoe}.
Attention provides a related setting in which Q/K/V projections shared across
positions can adapt to the mask. We examine how the number of trainable layers
affects this adaptation.

We present four controlled experiments on a 31M-parameter model
that isolate different aspects of the absorption mechanism
(Figures~\ref{fig:convergence} and~\ref{fig:absorption_gradient}),
with preliminary scale evidence from Qwen3-1.7B consistent with
the phenomenon persisting at 55$\times$ larger scale.
We connect the results to the MoE literature and discuss
implications for recent sparse attention methods that rely on
learned routing.

\section{Background and Setup}

\subsection{The Sparse Attention Routing Problem}

Given a pretrained transformer with attention scores
$A = QK^\top / \sqrt{d_h}$ and attention weights
$P = \text{softmax}(A)$, the sparse attention routing problem is:
learn a gate function $G(x)$ that predicts which entries of $P$
to keep, such that masking the rest preserves model quality.
The gate produces scores $G^{(h)} \in \mathbb{R}^{n \times n}$;
at deployment, only the top-$k$ entries per query are retained
and the rest are masked to $-\infty$ before softmax.

In the \emph{end-to-end} setting, $G$ is trained jointly with the
model: the gate and the Q/K/V projections co-evolve.
In the \emph{post-hoc} setting~\cite{ye2024seerattention}, the
model is frozen and only the gate is trained, typically by
distillation against the model's own attention distributions.

\subsection{Experimental Setup}
\label{sec:setup}

The 31M experiments use a 6-layer, 256-dimensional, 4-head pre-norm
transformer ($\sim$31M parameters) trained on WikiText-103
(512-token chunks, batch size 16, cosine LR schedule).
The dense baseline achieves 37.32 perplexity.

The gate adds per-head projections
$W_{gq}^{(h)}, W_{gk}^{(h)} \in \mathbb{R}^{d \times d_\text{gate}}$
producing gate scores
$G^{(h)} = (xW_{gq}^{(h)})(xW_{gk}^{(h)})^\top / \sqrt{d_\text{gate}}$.
With $d_\text{gate}$=32, this adds 393K parameters (1.3\% of the model).

We use this small model deliberately: it is large enough to exhibit
routing absorption but small enough to run end-to-end sparse training
experiments (50{,}000 steps) and controlled ablations at modest cost.
The full experimental suite (31M pretraining, all ablations, and
Qwen3 fine-tuning) cost under \$150 on rented
GPUs; every result in this paper can be independently reproduced
on a single consumer GPU in under a day.\footnote{Code and data:
\url{https://github.com/no-way-labs/routing-absorption}}
Section~\ref{sec:qwen3_absorption} examines adaptation in a larger model,
and Section~\ref{sec:scale} compares post-hoc sparsification across scales.

\subsection{Connection to MoE Routing Absorption}

Related effects have been reported in Mixture-of-Experts models.
Roller et al.~\cite{roller2021hash} showed that hash-based (random)
routing matches learned routing in MoE language models.
Chen et al.~\cite{chen2023smoe} and Fan et
al.~\cite{li2024empiricalmoe} reported results consistent with expert co-adaptation reducing the benefit
of learned routing.
Clark et al.~\cite{clark2022unified} provided scaling laws showing
that routing quality has diminishing returns as expert capacity grows.

A proposed explanation is that a large parameter budget in the experts allows
them to compensate for the routing decisions.
In attention, the full model ($\sim$31M parameters: Q/K/V projections,
feedforward layers, and embeddings) plays the role of experts, and
the gate ($\sim$393K parameters) plays the role of the router, a
roughly 80:1 parameter asymmetry that may facilitate compensation.

\section{Evidence for Routing Absorption}
\label{sec:evidence}

We present four experiments addressing gate performance, gradient flow,
mask dependence, and stochastic masking.
All use $k$=64 (87.5\% sparsity) unless otherwise noted.

\subsection{Experiment 1: Limited Benefit from Learned Soft Gates}
\label{sec:exp1}

We train the 31M model end-to-end with differentiable soft gating
for 50{,}000 steps.
The mask is applied as element-wise multiplication with sigmoid gate
scores (no hard top-$k$, so gradients flow through the gate).
We compare two conditions:

\begin{table}[t]
\centering
\begin{tabular}{lccc}
\toprule
Condition & Validation PPL & Seeds & Gate trainable? \\
\midrule
Learned soft gate  & 48.73 $\pm$ 0.60 & 3 & Yes (50K steps) \\
Random soft gate   & 49.83 $\pm$ 0.04 & 3 & No (frozen) \\
\midrule
Dense baseline     & 37.32 & N/A & N/A \\
\bottomrule
\end{tabular}
\caption{End-to-end soft gating at $k$=64.
Mean $\pm$ std over 3 seeds (42, 137, 256).}
\label{tab:softgate}
\end{table}

The gap is 1.10 ppl, or 2.2\% (Table~\ref{tab:softgate}).
Learned gates achieve \textbf{48.73 $\pm$ 0.60} ppl, compared with
\textbf{49.83 $\pm$ 0.04} for frozen random gates across three seeds.
The improvement is approximately 9\% of the gap between the random-gate
condition (49.83) and the dense baseline (37.32). Both gated conditions retain
substantially higher perplexity than the dense model. The small learned-gate
advantage is consistent with model adaptation reducing the contribution of the
routing function. The same gate architecture reaches near-oracle performance
when trained post-hoc on a frozen checkpoint in 1{,}000 steps
(Section~\ref{sec:posthoc}), indicating that the limited end-to-end benefit
depends on the training setting.

\paragraph{Convergence with frozen representations.}
When Q/K/V are frozen, the same gate architecture reduces perplexity from
46.86 to 37.33 in 500 steps, accounting for over 99\% of its final improvement
(Section~\ref{sec:gateonly}). This rapid convergence demonstrates that the gate
can learn useful routing from stable targets. Its limited benefit after
50{,}000 end-to-end steps is consistent with interference from co-adaptation;
longer joint-training budgets remain untested.

\subsection{Experiment 2: Gradient Flow through Hard Top-\texorpdfstring{$k$}{k}}
\label{sec:exp2}

Hard top-$k$ gating introduces a separate obstacle: the tested implementation
provides no task-gradient path to the gate scores.
The mask $M = \mathbf{1}[\text{rank}(G) \leq k]$ is
piecewise-constant with $\partial M / \partial G = 0$.
In PyTorch, the implementation chain
$\texttt{topk()} \to \texttt{scatter\_()} \to \texttt{masked\_fill()}$
produces a hard binary mask with no gradient path back to $G$.

Hard top-$k$ gating gives 71.22 ppl with nominally learned gates and 71.24 ppl
with frozen random gates, consistent with the absence of gate gradients.

Differentiable soft gating addresses this gradient-path limitation.
Experiment~1 nevertheless shows only a 2.2\% improvement over random gates under
soft gating. Together, the experiments identify two limitations of the tested
training procedures: hard masking blocks gate gradients, and enabling those
gradients leaves a substantial gap to dense-model quality under joint training.

\subsection{Experiment 3: The Distillation Contrast}
\label{sec:exp3}

We next test mask dependence by distilling gates on two checkpoints: one with mask-agnostic Q/K/V (dense-trained), one with
co-adapted Q/K/V (soft-gated, from Experiment~1).

In both cases, we freeze the model and train only the gate
projections for 1{,}000 steps using BCE loss against the oracle
top-$k$ mask.
Both gates converge to high F1 against their respective oracles.
Deployment with hard top-$k$ yields substantially different perplexities:

\begin{table}[t]
\centering
\begin{tabular}{lccc}
\toprule
Source checkpoint & Gate F1 & Deploy PPL & Interpretation \\
\midrule
Dense (mask-agnostic)  & 0.842 & 48.6  & Lower deployment PPL \\
Soft-gated (co-adapted) & 0.804 & 601.6 & Higher deployment PPL \\
\bottomrule
\end{tabular}
\caption{The distillation contrast ($k$=64).
Both gates predict oracle masks well (F1\,$>$\,0.8).
Hard-mask deployment yields a roughly 12$\times$ perplexity difference between
the checkpoints, despite similar mask-prediction F1.}
\label{tab:contrast}
\end{table}

The deployment difference in Table~\ref{tab:contrast} is consistent with
representations that depend on the training-time masking function. Replacing
sigmoid gating with binary top-$k$ changes both the selected entries and their
weighting. High F1 against the oracle mask is therefore insufficient to preserve
the co-adapted model's quality. The dense-trained checkpoint performs substantially
better under the same distillation and hard-mask deployment procedure,
supporting the use of representations learned independently of the gate.

\subsection{Experiment 4: Deployment after Stochastic Mask Training}
\label{sec:exp4}

A natural hypothesis is that co-adaptation can be prevented by
randomizing the mask during training, analogous to
dropout~\cite{hinton2012dropout} preventing feature co-adaptation.
We test this by training with a fresh random mask per forward pass
(a Gumbel-softmax attention-pattern dropout).

\begin{table}[t]
\centering
\begin{tabular}{lcc}
\toprule
Deployment condition & PPL & Notes \\
\midrule
Dense (no mask) & 78.19 & Mask removed \\
Fixed random masks (5 seeds) & 104.43 $\pm$ 0.66 & \\
Dense baseline & 37.32 & \\
\bottomrule
\end{tabular}
\caption{Stochastic mask training.
Dense deployment after stochastic mask training yields 78.19 ppl, compared with
37.32 for the dense-trained baseline. Fixed random-mask deployment gives a
further increase to $104.43 \pm 0.66$ ppl.}
\label{tab:stochastic}
\end{table}

The stochastically trained model retains more than twice the baseline
perplexity when evaluated without a mask (Table~\ref{tab:stochastic}). The tested
randomization procedure therefore fails to preserve dense-model quality. Its
flatter attention distributions are consistent with adaptation to mask noise
at the expense of concentrated attention patterns. These results limit the
usefulness of this particular masking scheme as a regularizer. Other stochastic
strategies, including annealed mask noise or mixtures of dense and sparse
forward passes, remain untested.

\section{Post-hoc Training with Frozen Representations}
\label{sec:posthoc}

Post-hoc training holds the model fixed while learning the routing function.
We examine the available attention structure and the resulting gate convergence.

\subsection{Concentration of Dense Attention}
\label{sec:structure}

\begin{figure}[t]
\centering
\includegraphics[width=0.55\linewidth]{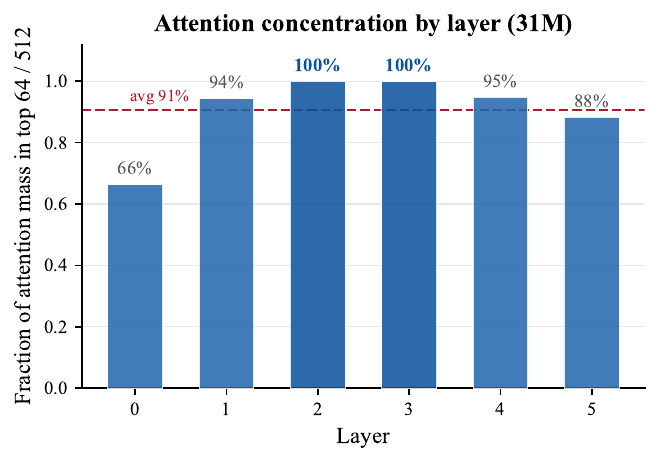}
\caption{Attention concentration by layer in the 31M model.
Each bar shows the fraction of total attention mass captured by the
64 highest-weight positions out of 512.
Layers~2 and~3 concentrate nearly all attention mass
in the top~64; even the least concentrated layer captures 66\%.
The dashed line shows the average across all layers (90.6\%).}
\label{fig:concentration}
\end{figure}

The dense model's attention distributions contain strong, learnable
structure (Figure~\ref{fig:concentration}).
Across all layers, the top 64 positions per query (out of 512)
capture 90.6\% of the total attention mass.
Some layers are even more concentrated: layer~2 concentrates
100\% of its attention mass in the top-64, and layer~3 captures
99.8\%.
The entropy ratio (actual entropy / maximum entropy) averages
0.497, well below the maximum-entropy value of 1.0 that would
indicate uniform attention.

The concentration measurements indicate substantial scope for sparsification.
Freezing Q/K/V allows the gate to learn these patterns from a stable target.

\subsection{Frozen Q/K/V Enable Fast Convergence}
\label{sec:frozen}

When the model is frozen, the gate faces a static optimization
landscape, with fixed attention-distribution targets for each input.
Table~\ref{tab:posthoc} reports the resulting distillation performance.

\begin{table}[t]
\centering
\begin{tabular}{lcccc}
\toprule
Condition & Steps & Gate F1 & PPL & vs.\ Oracle \\
\midrule
End-to-end learned gate  & 50{,}000 & N/A & 48.73 & +6\% \\
Post-hoc KL distillation & 1{,}000  & 0.833 & 48.83 & +6\% \\
Post-hoc KL ($k$=128)    & 1{,}000  & 0.888 & 40.24 & +2\% \\
Post-hoc KL ($k$=256)    & 1{,}000  & 0.931 & 37.57 & +0.2\% \\
\midrule
Oracle top-$k$ ($k$=64)  & N/A & 1.000 & 46.00 & N/A \\
Dense baseline            & N/A & N/A & 37.32 & N/A \\
\bottomrule
\end{tabular}
\caption{Post-hoc KL distillation on the frozen 31M model.
At $k$=64, 1{,}000 steps of post-hoc gate training yield 48.83 ppl,
compared with 48.73 for 50{,}000 steps of end-to-end training, using
50$\times$ fewer gate-training steps.
At moderate sparsity ($k$=256), perplexity is within 0.7\% of
the unmasked dense model.}
\label{tab:posthoc}
\end{table}

At $k$=64, post-hoc distillation reaches similar perplexity to end-to-end
training with 50$\times$ fewer gate-training steps (Table~\ref{tab:posthoc}).
Increasing $k$ further reduces post-hoc perplexity, reaching 37.57 at $k$=256.
These results support decoupling representation learning from gate training.
Freezing the dense checkpoint is the decoupling procedure tested here; the
performance of other procedures remains an empirical question.

\subsection{Gate-Only Training: Isolating the Gate}
\label{sec:gateonly}

To measure the gate's contribution with fixed representations, we run a
gate-only training experiment: freeze the dense checkpoint, train
\emph{only} the gate projections, and measure whether learned gates
improve over random.

\begin{table}[t]
\centering
\begin{tabular}{lcc}
\toprule
Condition & Final PPL & $\Delta$ PPL \\
\midrule
Learned gates (trainable)  & 37.29 $\pm$ 0.00 & $-$9.57 \\
Random gates (frozen)      & 46.86 $\pm$ 0.10 & 0.00 \\
\bottomrule
\end{tabular}
\caption{Gate-only training with frozen Q/K/V (3 seeds).
When co-adaptation is prevented by freezing, the gate drives
perplexity from 46.8 to 37.29 (near-dense), a 20\% improvement
relative to the frozen random-gate baseline.
The learned gate converges to the same reported perplexity across all seeds
($\sigma$=0.00 at the reported precision), while the random baseline varies only in its
initialization ($\sigma$=0.10).
Convergence is fast: over 99\% of the improvement occurs in 500 steps
(37.33 ppl), with the remaining 9{,}500 steps contributing only
0.04 ppl.}
\label{tab:gateonly}
\end{table}

\begin{figure*}[t]
\centering
\includegraphics[width=\linewidth]{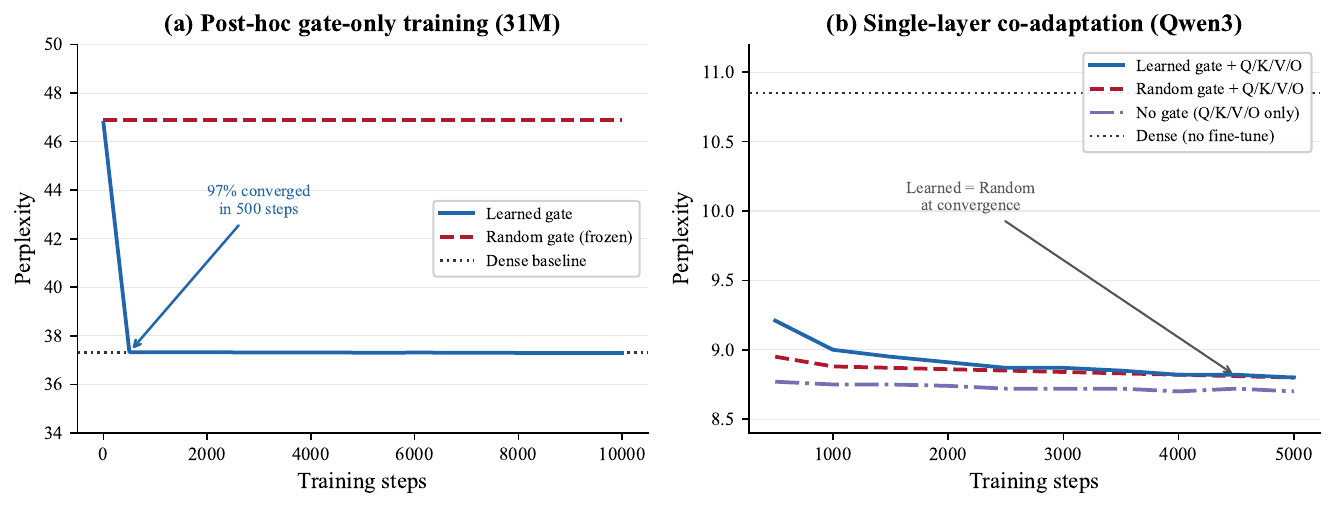}
\caption{Convergence dynamics under decoupled vs.\ co-adapted training.
\textbf{(a)}~Post-hoc gate-only training on the frozen 31M model:
the learned gate converges from 46.8 to 37.3 ppl in 500 steps
($>$99\% of total improvement), while the frozen random gate stays flat.
\textbf{(b)}~Single-layer co-adaptation at Qwen3 scale: learned and
random gates converge to identical perplexity (8.80), with the
no-gate control reaching 8.70. The panels show rapid gate learning with frozen
representations and no observed learned-gate advantage in the single-layer
co-adaptation experiment.}
\label{fig:convergence}
\end{figure*}

Across three seeds, learned gates converge to $37.29 \pm 0.00$ ppl, while
frozen random gates remain at $46.86 \pm 0.10$
(Table~\ref{tab:gateonly} and Figure~\ref{fig:convergence}a). The learned-gate
outcomes are consistent across the tested initializations at the reported
precision. Convergence is rapid: the gate reaches 37.33 ppl at step 500,
accounting for over 99\% of its total improvement, with the remaining
9{,}500 steps contributing 0.04 ppl. Together with Experiment~1, this result
shows that the same gate architecture learns useful routing much more readily
when the model representations are fixed.

\section{Parameter Asymmetry as a Proposed Mechanism}
\label{sec:mechanism}

Parameter asymmetry may help explain the limited contribution of the gate
during joint training. The model contains $\sim$31M trainable parameters,
compared with 393K in the gate, giving it approximately 80$\times$ as many
parameters that can adjust in response to the imposed mask.

MoE models also allocate substantially more parameters to experts than to routing.
In a typical MoE layer, the router has $d \times E$ parameters
(where $E$ is the number of experts, typically 8--64), while the
experts collectively have $E \times d_\text{ff} \times d$ parameters.
The ratio is $d_\text{ff}$:1, typically 4:1 to 16:1, depending on
the feedforward dimension.
Clark et al.~\cite{clark2022unified} showed that the benefit of
routing quality diminishes as expert capacity grows, consistent
with absorption increasing with the parameter asymmetry.

For the attention gate used here, the parameter ratio can be written explicitly.
The gate has $2 \cdot n_h \cdot d \cdot d_\text{gate}$ parameters
(two projections per head), while Q/K/V have
$3 \cdot d \cdot n_h \cdot d_h$ parameters.
The ratio is $3 d_h / (2 d_\text{gate})$, typically 3--6$\times$
per layer.
The feedforward layers, layer norms, and embeddings can also adjust during
joint training, bringing the effective ratio to $\sim$80:1 in our
31M model.

\paragraph{Gate capacity ablation.}
To assess gate capacity separately from model co-adaptation, we sweep
$d_\text{gate} \in \{32, 64, 128\}$ on frozen Qwen3-1.7B using BCE distillation
(Table~\ref{tab:gate_capacity}). Increasing capacity produces mixed results
across sparsity levels within the fixed 1{,}000-step budget. All tested BCE
configurations retain substantially higher perplexity than KL distillation at
$d_\text{gate}$=32. KL uses the full attention distribution as its target and
achieves 99.9\% efficiency at $k$=64 in this setting. The comparison indicates
that the distillation objective has a larger effect than the tested changes in
gate capacity. It does not directly measure the effect of gate capacity on
absorption during joint training.

\begin{table}[t]
\centering
\small
\begin{tabular}{lccc}
\toprule
$d_\text{gate}$ & $k$=64 PPL & $k$=128 PPL & $k$=256 PPL \\
\midrule
32 (KL)  & \textbf{12.24} & \textbf{11.72} & \textbf{11.56} \\
\midrule
32 (BCE)  & 102.82 & 25.76 & 20.78 \\
64 (BCE)  & 84.83  & 27.42 & 18.65 \\
128 (BCE) & 88.57  & 30.50 & 19.18 \\
\bottomrule
\end{tabular}
\caption{Gate capacity ablation on Qwen3-1.7B (post-hoc distillation).
Increasing $d_\text{gate}$ from 32 to 128 (4$\times$ more gate
parameters) produces mixed changes in BCE distillation perplexity.
KL distillation at $d_\text{gate}$=32 outperforms all BCE
configurations by 70--230$\times$ in excess perplexity above dense,
indicating a larger effect from the distillation objective in this comparison.
Dense: 11.52 ppl; oracle ($k$=64): 11.57 ppl.}
\label{tab:gate_capacity}
\end{table}

\paragraph{Testing adaptation in a larger model.}
The parameter-asymmetry hypothesis motivates measuring how gate performance
changes as more model parameters become trainable. Full end-to-end sparse
pretraining at 1.7B scale exceeds the scope of these experiments. We
fine-tune selected attention layers of Qwen3-1.7B to test whether adaptation
reduces the benefit of learned gating at this scale.

\subsection{Direct Test: Single-Layer Absorption at Qwen3 Scale}
\label{sec:qwen3_absorption}

To test routing absorption at scale without full pretraining, we
run a controlled experiment on Qwen3-1.7B: freeze the entire model
except one layer's attention projections (Q/K/V/O), add soft
bilinear gates to that layer, and fine-tune for 5{,}000 steps.
We compare learned vs.\ frozen random gates, mirroring
Experiment~1 (Section~\ref{sec:exp1}) at 55$\times$ larger scale.

Routing absorption predicts that adaptation in the unfrozen layer will reduce
the advantage of learned gates over random gates.
The frozen layers provide context, while the single unfrozen layer
tests whether co-adaptation occurs when a 1.7B-scale attention
mechanism can respond to a gate.

\paragraph{Results.}
Table~\ref{tab:qwen3_absorption} shows the results.
The learned and random gates converge to identical perplexity
(8.80 vs.\ 8.80), consistent with routing absorption at 1.7B scale.
The no-gate control (fine-tuning only Q/K/V/O with standard
dense attention) reaches 8.70, giving both gated conditions a 0.10 ppl penalty
relative to this control.
The random gate also converges faster than the learned gate
at every checkpoint (e.g., 8.95 vs.\ 9.21 at step 500),
showing no convergence-speed advantage from learning the gate in this run.

\begin{table}[h]
\centering
\small
\caption{Single-layer absorption at Qwen3 scale (layer 14).
All conditions fine-tune the same Q/K/V/O projections for 5{,}000
steps with identical hyperparameters; only the gate differs.}
\label{tab:qwen3_absorption}
\begin{tabular}{lc}
\toprule
\textbf{Condition} & \textbf{PPL} \\
\midrule
Dense baseline (no fine-tune) & 10.85 \\
No gate (Q/K/V/O only) & 8.70 \\
Learned gate + Q/K/V/O & 8.80 \\
Random gate + Q/K/V/O & 8.80 \\
\midrule
Gate gap (learned $-$ random) & 0.00 \\
Gate overhead (gated $-$ no gate) & +0.10 \\
\bottomrule
\end{tabular}
\end{table}

\paragraph{Absorption gradient: varying co-adaptation capacity.}
The single-layer result motivates testing how the learned-gate advantage
changes with the number of trainable layers.
We test this by sweeping the number of unfrozen layers
$n \in \{0, 2, 4, 8\}$ (out of 28), with soft bilinear gates on all
layers, training for 5{,}000 steps at $k$=64.
When $n$=0, all Q/K/V are frozen and only the gates train
(the post-hoc setting); as $n$ increases, more layers can co-adapt.

\begin{table}[h]
\centering
\small
\caption{Absorption gradient at Qwen3 scale.
As more layers unfreeze, the random gate improves (42$\to$17 ppl)
while the learned-gate condition remains near 10 ppl.
The gap shrinks from 31.6 to 6.9 as more layers become available for adaptation.}
\label{tab:absorption_gradient}
\begin{tabular}{rcccc}
\toprule
$n_\text{unfrozen}$ & Learned PPL & Random PPL & Gap & Co-adapt.\ capacity \\
\midrule
0  & 10.71 & 42.26 & $-$31.55 & None (post-hoc) \\
2  & 10.13 & 23.07 & $-$12.94 & 7\% of layers \\
4  & 10.03 & 21.04 & $-$11.01 & 14\% of layers \\
8  & 10.48 & 17.40 & $-$6.92  & 29\% of layers \\
\bottomrule
\end{tabular}
\end{table}

\begin{figure}[t]
\centering
\includegraphics[width=0.75\linewidth]{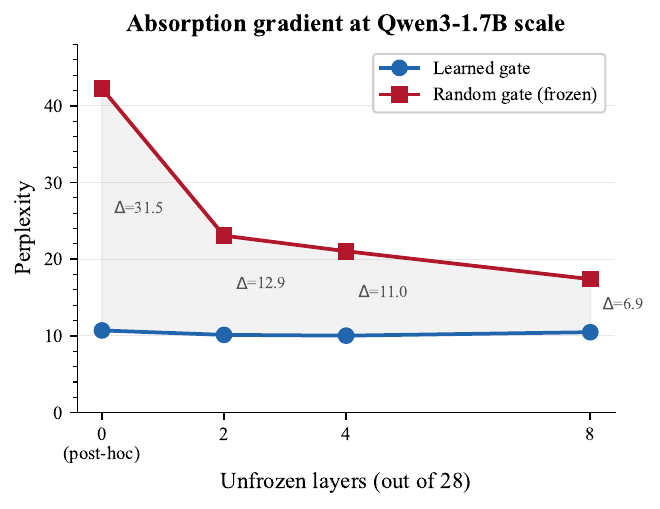}
\caption{The absorption gradient at Qwen3-1.7B scale.
As more layers unfreeze (increasing co-adaptation capacity),
the random gate's perplexity drops toward the learned gate's level.
The shaded area shows the gap shrinking from 31.5 (post-hoc, no
co-adaptation) to 6.9 (29\% of layers unfrozen).
The learned-gate condition remains near 10 ppl, so the narrowing gap
primarily reflects improvement in the random-gate condition.}
\label{fig:absorption_gradient}
\end{figure}

Table~\ref{tab:absorption_gradient} and
Figure~\ref{fig:absorption_gradient} show a clear gradient.
With all layers frozen ($n$=0), learned gates achieve 10.71 ppl
while random gates remain at 42.26, a 31.6 ppl gap confirming
that the gate \emph{can} learn useful routing when decoupled.

As layers unfreeze, the random gate's perplexity drops sharply:
consistent with Q/K/V adaptation compensating for the random mask.
At $n$=8 (29\% of layers), the gap has shrunk to 6.9 ppl,
closing 78\% of the original 31.6~ppl gap.
The closure rate is decelerating: the first 2 unfrozen layers
(7\% of the model) close 59\% of the gap, while the next 6 layers
close only 19\% more.
This is consistent with Experiment~1's endpoint at 31M scale,
where full end-to-end training drives the gap to 2.2\%.

The learned-gate condition remains near 10 ppl across the tested configurations,
while the random-gate condition improves as more layers become trainable.
This pattern supports compensation for random routing through model adaptation.
Perplexity alone leaves the extent to which the learned routing patterns change
across configurations unresolved.

\subsection{Post-hoc Sparsification across Model Scales}
\label{sec:scale}

The Qwen3 experiments demonstrate adaptation in a larger pretrained model.
A complementary comparison evaluates post-hoc distillation and oracle masking
at both scales, measuring how much quality is preserved when attention entries
are removed.

\begin{table}[t]
\centering
\begin{tabular}{lcccc}
\toprule
& \multicolumn{2}{c}{31M model} & \multicolumn{2}{c}{Qwen3-1.7B} \\
\cmidrule(lr){2-3} \cmidrule(lr){4-5}
$k$ & Oracle / Dense & Efficiency & Oracle / Dense & Efficiency \\
\midrule
64  & 46.0 / 37.3 = 1.23 & 94\% & 11.57 / 11.52 = 1.004 & 99.9\% \\
128 & 39.5 / 37.3 = 1.06 & 96\% & 11.53 / 11.52 = 1.001 & 99.9\% \\
256 & 37.5 / 37.3 = 1.01 & 99\% & 11.52 / 11.52 = 1.000 & 99.8\% \\
\bottomrule
\end{tabular}
\caption{Post-hoc distillation efficiency across scales.
At Qwen3 scale, oracle top-$k$ is nearly lossless even at $k$=64,
and the gate captures this almost perfectly ($>$99.8\% efficiency).
At 31M scale, attention is less concentrated (oracle $1.23\times$
dense at $k$=64) and the gate is correspondingly less efficient (94\%).}
\label{tab:scale}
\end{table}

\begin{figure*}[t]
\centering
\includegraphics[width=\linewidth]{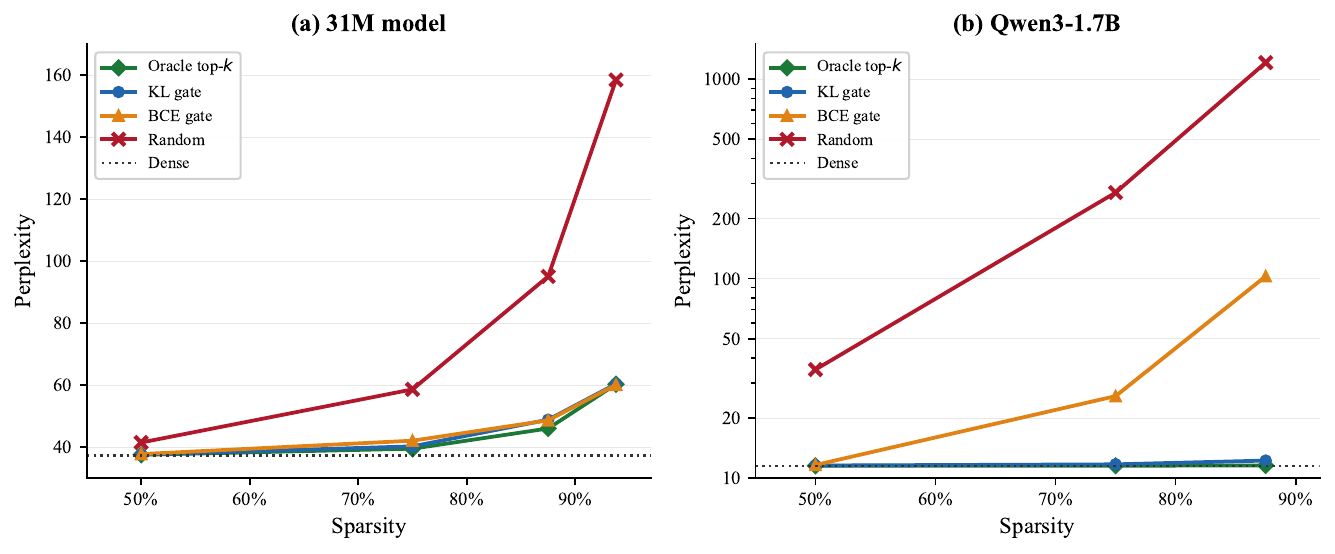}
\caption{Perplexity vs.\ sparsity for oracle, KL-distilled gate,
BCE-distilled gate, and random masking at two scales.
\textbf{(a)}~On the 31M model, KL and BCE gates perform similarly,
both tracking oracle closely.
\textbf{(b)}~On Qwen3-1.7B (log scale), KL distillation tracks
oracle at all sparsity levels while BCE yields substantially higher perplexity
at high sparsity (102.8 vs.\ 12.2 at 87.5\% sparsity).
The difference supports using the full attention distribution as the
distillation target in this setting.}
\label{fig:sparsity_ppl}
\end{figure*}

Oracle sparsification preserves more of the dense-model quality in Qwen3 than
in the 31M model (Table~\ref{tab:scale} and Figure~\ref{fig:sparsity_ppl}).
At $k$=64, Qwen3's oracle perplexity is within 0.4\% of dense, while the 31M
model's oracle perplexity is 23\% higher. These results indicate greater
post-hoc sparsification tolerance in the tested Qwen3 checkpoint. Sharper
attention may also increase sensitivity to gate perturbations during joint
training, providing a possible source of pressure for compensation. Establishing
how this effect varies with scale requires experiments that control for model
architecture and training history.

\section{Implications}
\label{sec:implications}

\subsection{For Sparse Attention Methods}

Several recent methods learn sparse attention patterns
end-to-end~\cite{lou2024sparseK,yuan2025nsa,piekos2025mosa}.
Our results motivate explicit measurement of the learned router's contribution
in per-query token-level masking methods. Q/K/V adaptation can reduce the
performance difference between learned and random routing, as observed in the
tested configurations.

We note that routing absorption as characterized here is specific
to \emph{token-level gated masking}, where a small gate produces
per-query top-$k$ masks over individual key positions.
Other sparse attention formulations may sidestep absorption through
different structural choices.
MoSA~\cite{piekos2025mosa} uses expert-choice routing over tokens
(selecting which tokens each head attends to), which is
architecturally distinct from per-query masking.
NSA~\cite{yuan2025nsa} operates at block granularity with a
compression branch.
These structural differences may alter the extent of co-adaptation. Related
variation across MoE formulations~\cite{clark2022unified} reinforces the need
for method-specific evaluation.

Ablations replacing learned routing with fixed or random
routing, in the style of Roller et al.~\cite{roller2021hash},
would be informative for any method claiming learned routing
provides a benefit.
For example, NSA combines compressed attention,
sliding windows, and learned sparse selection.
Replacing learned selection with random selection while retaining the windows
and compression would isolate the incremental benefit of the selection
component.

\subsection{For the MoE Literature}

Attention provides a complementary setting for studying router--model
co-adaptation. Oracle masks, distillation across checkpoints, and mask-swap
evaluations allow routing accuracy and deployment quality to be measured
separately. The roughly 12$\times$ deployment-perplexity difference in
Section~\ref{sec:exp3}, despite similar gate F1, illustrates why both measures
are needed.

\paragraph{Structural differences from MoE.}
MoE routers select among expert modules with separate parameters. Attention
gates select key positions whose representations are computed using shared
Q/K/V projections. This shared structure allows changes to the projections to
affect many routed key-value pairs simultaneously. The layer-unfreezing
experiment (Table~\ref{tab:absorption_gradient}) shows that additional trainable
attention layers improve performance under random gating: the learned-gate
advantage falls from 31.6 to 6.9 ppl when eight of 28 layers are unfrozen.
These results support compensation through attention-layer adaptation, while
the specific paths of information redistribution remain to be identified.
The experiments provide no direct comparison of co-adaptation severity between
attention and MoE.

ReMoE~\cite{shi2025remoe} uses ReLU routing to make MoE gating differentiable.
Our soft-gating experiment indicates that differentiability alone is insufficient
to prevent absorption in the tested attention model. Whether a related effect
occurs in ReMoE requires a separate comparison with fixed or random routing;
its other design choices, including variable-rate computation, may also affect
performance.

\subsection{The Decoupling Principle}

For the gates studied here, separating representation learning from routing
training improves the gate's contribution. This finding motivates examining
related uses of fixed targets and independently trained representations:

\begin{itemize}
\item \textbf{Post-hoc pruning.}
  Random pruning can preserve model quality in some
  settings~\cite{liu2022random,gadhikar2023random}. This motivates testing
  whether representations learned before sparsification are less dependent on
  particular masks, as observed for dense Q/K/V here.
\item \textbf{Knowledge distillation.}
  Training a small student against a frozen teacher's soft
  outputs~\cite{hinton2015distilling} provides a fixed target for student
  optimization. Joint teacher--student training would change that target and
  could introduce co-adaptation, a hypothesis requiring separate tests.
\item \textbf{Feature dropout.}
  Dropout~\cite{hinton2012dropout} prevents feature co-adaptation
  by randomly removing features during training.
  Our Experiment~4 shows that the analogous strategy for
  attention masks yields substantially higher deployment perplexity than dense
  training in the tested configuration. This result motivates evaluating mask
  regularization separately from feature dropout.
\end{itemize}

\section{Limitations}

Our controlled experiments use a 31M-parameter model for the
full end-to-end training ablations.
The single-layer Qwen3 experiment
(Section~\ref{sec:qwen3_absorption}) is consistent with absorption
at 1.7B scale but in a constrained setting (one unfrozen layer); full
end-to-end sparse pretraining at 1.7B scale remains out of scope.
The absorption gradient experiment
(Table~\ref{tab:absorption_gradient}) sweeps up to $n$=8 unfrozen
layers (29\% of 28).
At this point, 78\% of the gap between post-hoc and co-adapted
settings has already closed, with diminishing marginal returns
(the first 2 layers close 59\%; the next 6 close only 19\% more).
Even an experiment with all 28 layers unfrozen would remain a
5{,}000-step fine-tuning study, leaving full-pretraining behavior unresolved.
The 31M experiment (Section~\ref{sec:exp1}) allows all parameters to co-adapt
over 50{,}000 pretraining steps and yields a 2.2\% learned-gate advantage.
The Qwen3 experiments establish that a related reduction in the learned-gate
advantage occurs at 55$\times$ larger model scale under partial fine-tuning.

The 50{,}000-step soft-gating experiment tests one training duration. Longer
training could change the learned-gate advantage. Convergence within 500 steps
with frozen Q/K/V demonstrates that stable-target routing is readily learnable,
but leaves the convergence time of joint training unresolved. Parameter
asymmetry is a proposed explanation for the observed adaptation; the experiments
do not establish a general scaling law relating the parameter ratio to absorption.

We study only the bilinear gate form $g_q g_k^\top$.
More complex routing mechanisms (e.g., cross-attention between
queries and a compressed key representation) might resist
absorption better, though the MoE literature suggests that
router architecture matters less than the fundamental parameter
asymmetry~\cite{clark2022unified}.

Our post-hoc experiments use attention distributions and oracle masks as
distillation targets, requiring access to the model's attention scores.
Whether other forms of decoupled training (e.g., training the gate
on a separate dataset, or using a different teacher signal) work
equally well is untested.

\section{Conclusion}

In the tested 31M transformer, learned soft gates improve perplexity by 2.2\%
over frozen random gates after 50{,}000 steps of joint training. Gradient-path
checks, distillation across checkpoints, and stochastic-mask experiments identify
limitations in gate training and sensitivity to the masking function. Qwen3-1.7B
experiments further show that allowing more attention layers to adapt reduces
the performance gap between learned and random gates.

These observations support routing absorption, in which model adaptation reduces
the incremental benefit of learned routing. Parameter asymmetry provides a
possible explanation and connects the findings to co-adaptation in MoE.
Freezing the model enables rapid gate learning from stable attention targets
and effective post-hoc sparsification in the tested settings.

For per-query token-level gates, the results support post-training sparsification
and routine comparison with frozen random routing. Measuring both routing
accuracy and deployment perplexity helps distinguish the gate's contribution
from the model's adaptation to its mask.

\end{document}